# VARENN: Graphical representation of spatiotemporal data and application to climate studies

Takeshi Ise[1]* and Yurika Oba[1,2]

[1]Field Science Education and Research Center (FSERC), Kyoto University, Kyoto, Japan
[2]Center for the Promotion of Interdisciplinary Education and Research, Kyoto University, Japan
*corresponding author: ise@kais.kyoto-u.ac.jp

## Abstract

Analyzing and utilizing spatiotemporal big data are essential for studies concerning climate change. However, such data are not fully integrated into climate models owing to limitations in statistical frameworks. Herein, we employ VARENN (visually augmented representation of environment for neural networks) to efficiently summarize monthly observations of climate data for 1901–2016 into 2-dimensional graphical images. Using red, green, and blue channels of color images, three different variables are simultaneously represented in a single image. For global datasets, models were trained via convolutional neural networks. These models successfully classified rises and falls in temperature and precipitation. Moreover, similarities between the input and target variables were observed to have a significant effect on model accuracy. The input variables had both seasonal and interannual variations, whose importance was quantified for model efficacy. VARENN is thus an effective method to summarize spatiotemporal data objectively and accurately.

## Introduction

In the era of big data, spatiotemporal digital data, such as satellite observations, have been continuously accumulating [1]. Such data should be used to better understand and predict the behavior of the Earth system: from ecosystem to the climate [2]. However, these data are not fully utilized currently, possibly owing to the lack of a suitable framework to extract important information from such data [1, 3, 4]. Scientists work eagerly to improve simulation models for climatic prediction, but the performance of the models has not be sufficient. Specifically, interannual to decadal climatic patterns such as ENSO (El Nino/southern oscillation), NAO (North Atlantic oscillation), and PDO (Pacific decadal oscillation) are not well reproduced by current climate models (i.e., Earth system models), indicating that the models are not fully optimized with the available data [5, 6], especially for decadal teleconnection patterns.

Deep learning has been used in diverse topics recently. For example, CNNs (convolutional neural networks) have been a powerful tool in computer vision for object identification and classification in various fields [7]. Related to Earth system science, computer vision has been applied for satellite-based images [8, 9] and field-based pictures [4]. Using visual images, a CNN model successfully detected extreme weather conditions [10]. However, these studies mainly use 2-dimensional image data. Because targets for Earth system modeling tend to have a dimension in time, a newer approach that explicitly integrates temporal dynamics is required [1]. For instance, convolutional LSTM [11] is one of the methods to explicitly treat spatiotemporal data. Earth system data are characterized by the large size (i.e., big data) with complex spatiotemporal interactions such as seasonal and interannual climate variability. Ise and Oba [12] approached such data by converting global temperature data into artificial 2-dimensional images and demonstrated that the resultant model successfully summarized temporal characteristics globally, as a classification problem. We believe that this approach should be augmented and used with various data from atmospheric, oceanic, and terrestrial ecosystems.

In this study, we propose a novel data processing method called VARENN (visually augmented representation of environment for neural networks) and apply it to global time series of 8 climatic variables from 1901–2016 to construct a classification model via supervised learning with a CNN. A few previous studies have reported methods of converting time series data into images and performed analyses using CNNs, but they only utilized one variable [12, 13]. Here, by using the three (red, green, and blue) channels of digital images, we integrate three different time-series variables into a single image. By doing this, we aim to classify decadal interaction patterns of climatic variables. We design a series of experiments to demonstrate that VARENN is an effective method to integrate environmental variables into common CNN frameworks.

# Results

VARENN is designed to convert time series data containing both seasonal and interannual variabilities into 2-dimensional images with artificial color assignment. Because a color image is composed of 2-dimensional arrays of three color channels, it is possible to assign three different time series to such arrays. This facilitates the analysis of multiple global time series data integratively via CNN. In this study, using multiple climatic variables for global terrestrial regions, we attempted to construct models to classify future rises and falls in temperature and precipitation, according to information from the multiple time series.

In this study, 8 spatiotemporal climatic variables were obtained from the Climatic Research Unit Time-Series v. 4.01 (CRU TS4.01, Table 1) [14, 15]. CRU TS4.01 is a high-resolution gridded dataset for terrestrial areas divided into in 0.5° grids. It contains monthly climatic data for the period of 1901–2016. From the eight climatic variables, we systematically selected up to three variables and assigned them to color channels (RGB) to construct artificial color images conveying varied climatic information.

To visualize seasonal and interannual variations and trends, a 30-year window (training period: duration of time where we obtained data for VARENN images) was systematically selected from the 116 years of the dataset. We created graphical images of 60×60 pixels from the 30-year window data (Fig. 1). The climatic variables were scaled to 0–1 to fit the color spaces. Monthly data for a given year were vertically aligned from the top, and the yearly data were horizontally aligned from the left.

To systematically analyze the model performance and find the optimal combination of climatic variables, 92 combinatorial experiments were executed (Supplementary Table 1): eight models used one climatic variable (1-VAR experiments), 28 models used two climatic variables (2-VAR experiments), and 56 models used three climatic variables (3-VAR experiments). This combinatorial analysis was conducted independently for temperature change experiments (TMP-EX) and precipitation change experiments (PRE-EX). In contrast to the previous study with one variable [12], here we superimpose up to three different climatic variables into color (RGB) channels. For 1-VAR models, the selected climatic variable was allocated to the R channel. For 2-VAR models, the R and G channels were used. The vacant channel(s) was/were filled with zeroes. For 3-VAR models, the R, G, and B channels were all used.

The images are annotated with five categories based on decadal climatic trends according to the following relationships:

$$F_T = \begin{cases} T_1, & 5 \le \mu_{10} - \mu_{30} \\ T_2, & 2.5 \le \mu_{10} - \mu_{30} < 5 \\ T_3, & 0 \le \mu_{10} - \mu_{30} < 2.5 \\ T_4, & -2.5 \le \mu_{10} - \mu_{30} < 0 \\ T_5, & \mu_{10} - \mu_{30} < -2.5 \end{cases}$$

$$F_P = \begin{cases} P_1, & 30 \le \mu_{10} - \mu_{30} \\ P_2, & 10 \le \mu_{10} - \mu_{30} < 30 \\ P_3, & -10 \le \mu_{10} - \mu_{30} < 10 \\ P_4, & -30 \le \mu_{10} - \mu_{30} < -10 \\ P_5, & \mu_{10} - \mu_{30} < -30 \end{cases}$$

where $F_T$ and $F_P$ are the classification categories for TMP-EX and PRE-EX, respectively, $\mu_{30}$ is the mean of the target variable (tmp or pre) for the 30-year training period, and $\mu_{10}$ is the mean of the target variable for the labeling period (a 10-year window following the training period). For TMP-EX, the unit is °C. For PRE-EX, the unit is mm. $F_T$ or $F_P$ categories were assigned as annotation for supervised learning.

We conducted 92 combinatorial experiments for both TMP-EX and PRE-EX (Fig. 2; Supplementary Table 1) with randomly chosen images ~100,000. The test accuracy tended to be high when the target variable (tmp or pre) was in the input variables. The test accuracy considerably varied among experiments with different input variables. For instance, in experiments for 3-VAR TMP-EX, the highest test accuracy (0.829) was achieved with pet, tmp, and vap, whereas the lowest accuracy (0.739) was achieved with frs, pre, and wet.

Using the non-parametric Kruskal–Wallis test, we observe that there are significant differences among the accuracies of the 1-VAR, 2-VAR, and 3-VAR experiments ($p<0.05$) for both TMP-EX and PRE-EX. The result shows that the number of climatic variables had an obvious effect on the accuracy of the classification model. Then, with the Mann–Whitney *U* test, we check the differences in test accuracies for each pair of the three experiments. In TMP-EX, the accuracy differences of 1-VAR and 3-VAR were significant ($p<0.05$). In PRE-EX, the accuracy differences of 1-VAR, 2-VAR, and 3-VAR were significant ($p<0.05$).

Because the eight climatic variables used in this study have varying levels of similarity (Supplementary Fig. 1), we examine the effects of the relationships between the target variables (i.e., tmp or pre) and input variables (i.e., cld, dtr, frs, pet, pre, tmp, vap, and wet). To quantify the similarity among variables, we calculated the Euclidean distances from each target variable to the input variable. Then, we analyzed the relationship between the similarity of variables and test accuracy by using linear regression. As a result, we found that the accuracy of models with variables that are closely related to the target tend to be higher. The regression lines in Fig. 2 are all statistically significant ($p<0.05$) except 1-VAR PRE-EX. The eight climatic variables used in this study have varied levels of similarity (Supplementary Fig.

1), and we examine the effects of the relationships among the variables that were assigned to R, G, and B to the model performance.

The best model in 3-VAR was the combination of pet, tmp, and vap for TMP-EX (#86 in Supplementary Table 1), and the combination of frs, pre, and wet for PRE-EX (#48). To maximize the performance, we increased the number of training images to ~1.5 million. For these high-performance models of TMP-EX #86 and PRE-EX #48, resultant test accuracies were 0.937 and 0.828, and the weighted Kappa statistics were 0.959 and 0.867, respectively. These results suggest that the VARENN framework is scalable; the greater the amount of training data, the higher the accuracy. Using these high-performance models, we visualized the model classifications and errors (Fig. 3). Basically, the models were able to assign correct categories globally. However, classification errors were concentrated on the border of $F_T$ or $F_P$ categories. This is reasonable because VARENN images of border grids tend to contain characteristics of both categories across the border. It should be noted that the partitioning of categories are arbitrary; if classification schemes other than $F_T$ or $F_P$ were used, the erroneous borders would change positions. Comparing TMP-EX and PRE-EX, it should be noted that the distribution patterns of 5 categories are more complex for the latter. This would increase the length of borders and result in lower accuracies found in PRE-EX than TMP-EX.

To check if seasonal and/or interannual variations are critical for constructing an appropriate model, we designed "knockout" experiments, in which we intentionally "inactivate" seasonal or interannual variations (Table 2). When we remove interannual variations (i.e., monthly-averaged climatic variables for 30 years), the VARENN images resemble squares with horizontal stripes. On the other hand, when we remove seasonal variations (i.e., annual average climatic variables), the images have vertical stripes. When we construct models with these images, the reduction in information lowered test accuracies. This experiment strongly suggests that both seasonal and interannual variabilities are critical for constructing appropriate models. In addition, to quantify the effect of the length of training period, we carried out experiments with training period of 10 years and compared test accuracies with the 30-year experiments (Table 2). The reduction in the duration of the training period lowered model accuracy considerably, indicating that the multi-decadal climate patterns in the 30-year training period convey information to classify climatic patterns more accurately.

# Conclusions

In the series of experiments, we successfully illustrated that VARENN can effectively represent global time series data of climate, with seasonal and interannual variations. Applying this method to gridded data, we were able to classify climatic patterns in the time series. We examined 92 combinations of variables for TMP-EX and PRE-EX. As a result, the number of climatic variables had a clear effect on the accuracy of the classification model; the greater the number of variables embedded in training images, the higher the test accuracy obtained. We also found that the model performance is related to the similarity between the target and input variables.

Our approach, VARENN, has some similarities with convolutional LSTM [11]. Both approaches are designed to explicitly treat spatiotemporal data. However, the former has several unique characteristics: (1) two time-series trends (i.e., seasonal and interannual) are graphically represented, (2) arbitrary combinations of multiple variables are accepted, and (3) global big data, instead of geographically limited data, can be fed to the model. Thus we suggest that VARENN can be an effective analysis tool for environmental and earth science.

The conversion of times-series signals into artificial, 2-dimensional images to detect anomalies using CNN has been suggested previously [13]. However, the signals used in that study did not have well-defined periodicity such as seasonality and thus the dimensions of the image were set arbitrarily. In this study, in contrast, we have set the dimensions according to the explicit seasonality and thus VARENN images exhibit clear patterns. Superimposed multiple signals in RGB color channels are another unique characteristic of VARENN. Thus we believe that our approach are novel, especially for large-scale spatiotemporal environmental data.

Process-based simulation models will remain as the core for projections of the climate because statistical methods including AI have limited ability to predict the future in novel conditions where it is difficult to obtain training data. We suggest using AI-based methods such as VARENN to improve process-based models. For example, in this study, the model successfully classified interannual to decadal climatic patterns. We selected the window size (30 year for training and 10 year for labeling) to capture the temperature and precipitation patterns occurring in timescale of teleconnection patterns such as ENSO. By comparing the results of AI-based (statistical) studies and process-based (physical) simulations, the accuracy and precision of the simulations can be evaluated appropriately. Objective parameterization such as data assimilation of process-based simulation models are suggested [3, 16], but feature extraction with deep learning is unique because it proposes model itself, not just parameters, objectively.

The VARENN approach aims to find climatic patterns as a classification problem. We believe that our approach has successfully captured multi-decadal climatic signatures because the experiments with 30-year training period had better performance than 10-year experiments. Our study should be augmented in the near future. For example, by using a visualization technique, it is possible to obtain visual explanations for the classification done by the model. By doing this, we may determine which part of the image is important for the AI decision, whether such decisions differ among different climatic regions, how much time lag is considered, and if a pattern can be identified for extreme weather events. In summary, the method to analyze spatiotemporal big data will be a promising tool for improving studies and projections concerning climate change.

# Methods

The hardware used to run and test the CNN had an Intel Core i7-9700K CPU, 32 GB RAM, and a NVIDIA GeForce RTX 2080 Ti GPU and the operating system was Ubuntu 16.04 LTS. The backend for the CNN was TensorFlow 1.12.0 implemented on NVIDIA DIGITS 6.1 (provided as NVIDIA GPU Cloud image DIGITS Release 18.12) [17]. We employed a CNN with LeNet (Supplementary Fig. 2) [18]. The number of training epochs was 30, and the solver type was stochastic gradient descent (SGD). We constructed all images in 60×60 squares to maximize efficiency of DIGITS 6.1. One of the advantages of NVIDIA DIGITS 6.1 was the easy implementation of gradually changing learning rates, which was utilized when we constructed the high-performance models with ~1.5 million images. In our environment, the time consumed for learning was ~5 minutes for ~100,000 images and ~1 hour for ~1.5 million images. With the test datasets, we constructed confusion matrices and calculated the test accuracy. We also calculated the weighted Kappa statistic for these high-performance models. Relationships among 1-VAR, 2-VAR, and 3-VAR experiments were analyzed by non-parametric significance test of Kruskal–Wallis test and Mann–Whitney $U$ test with Bonferroni adjustment. VARENN image generation and statistical analyses were performed using R 3.5.1 [19].

**Selecting Images and Supervised Learning by CNN**

Globally, there are ~67,000 terrestrial grid cells with a resolution of 0.5°. The length of the time series is 116 years. When we systematically shift the window of 40 years (30 years for training periods and 10 years for labeling periods), there can be 116 – 40 + 1= 77 images in one grid. Thus, the maximum number of images is 67,000 × 77 ≈ 5,200,000. However, owing to limitations in computational resources, using the whole images would significantly deter the study. Thus, we decided to reduce the number of images to increase the efficiency of the study design. We assigned a random number $c$ from 0 to 1 for each grid. When $c > c_t$, where $c_t$ is the threshold, the grid is selected for Image Set, which is the dataset used in the experiments. By changing $c_t$, we could balance the efficiency and model performance. For example, to perform a 2×92 combinatorial analysis, we set $c_t$=0.98. This limited the number images in Image Set to ~100,000. When we constructed high-performance models, we set $c_t$=0.7 to use ~1.5 million images. By using the built-in exponential decay function in DIGITS 6.1 to prevent excessive overfitting, we gradually reduced the learning rates as epoch progressed. Image Set was partitioned into training, validation, and test subsets in portions of 75%, 20%, and 5%, respectively. Because our study concerns time series, we assign the label training, validation, or test to each grid, rather than each image, to prevent partial duplications of images, which may induce overestimation of accuracy, in the 3 subsets. Selected DIGITS 6.1 outputs for the supervised learning are shown in Supplementary Fig. 3.

## Acknowledgements

This work was supported by JSPS KAKENHI Grant Number JP18H03357 and the Link Again Program of the Nippon Foundation-Kyoto University Joint Project.

## Author Contributions

TI prepared environments for the study. YO performed experiments and summarized the results. TI and YO contributed equally.

## Competing Interests

The authors declare no conflicts of interest.

# Figure legends

**Fig. 1**. Examples of VARENN images (a) for TMP-EX with pet, tmp, and vap assigned to R, G, and B channels, respectively, and (b) for PRE-EX with cld, pet, and pre assigned to R, G, and B channels, respectively.

**Fig. 2**. Plots to compare performances of 1-VAR, 2-VAR, and 3-VAR experiments. (a) Summary for TMP-EX. (b) Summary for PRE-EX.

**Fig. 3**. Visualization of classification by high-performance models. (a) TMP-EX #86. (b) PRE-EX #48. (c) Classification errors (indicated in orange) for TMP-EX #86. (d) Classification errors for PRE-EX #48.

# Tables

**Table 1**. Eight climatic variables used in this study, from CRU TS 4.01. All data are prepared as monthly datasets for 1901–2016, and the spatial resolution is 0.5°.

| **label** | **variable** | **units** |
|---|---|---|
| cld | cloud cover | % |
| dtr | diurnal temperature range | °C |
| frs | frost day frequency | days |
| pet | potential evapotranspiration | mm $d^{-1}$ |
| pre | precipitation | mm $mo^{-1}$ |
| tmp | daily mean temperature | °C |
| vap | vapor pressure | hPa |
| wet | wet day frequency | days |

**Table 2**. Test accuracy of models of “knockout” experiments of temporal variations and “shortened” experiments.

| experiments | sample images | test accuracies | |
|---|---|---|---|
| | | TMP-EX#86 | PRE-EX#48 |
| Default (with both seasonal and interannual variations, 30-year training) | 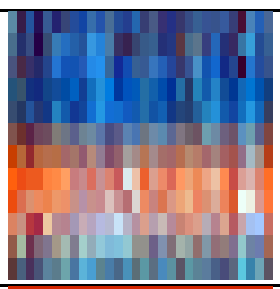 | 0.829 | 0.638 |
| Seasonal variations only | 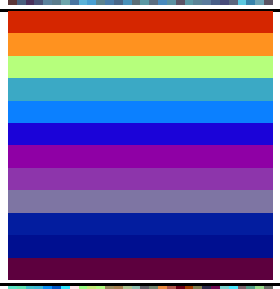 | 0.569 | 0.489 |
| Interannual variations only | 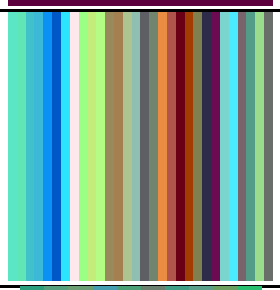 | 0.766 | 0.517 |
| 10-year training | 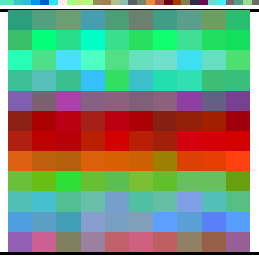 | 0.742 | 0.562 |

# Figures

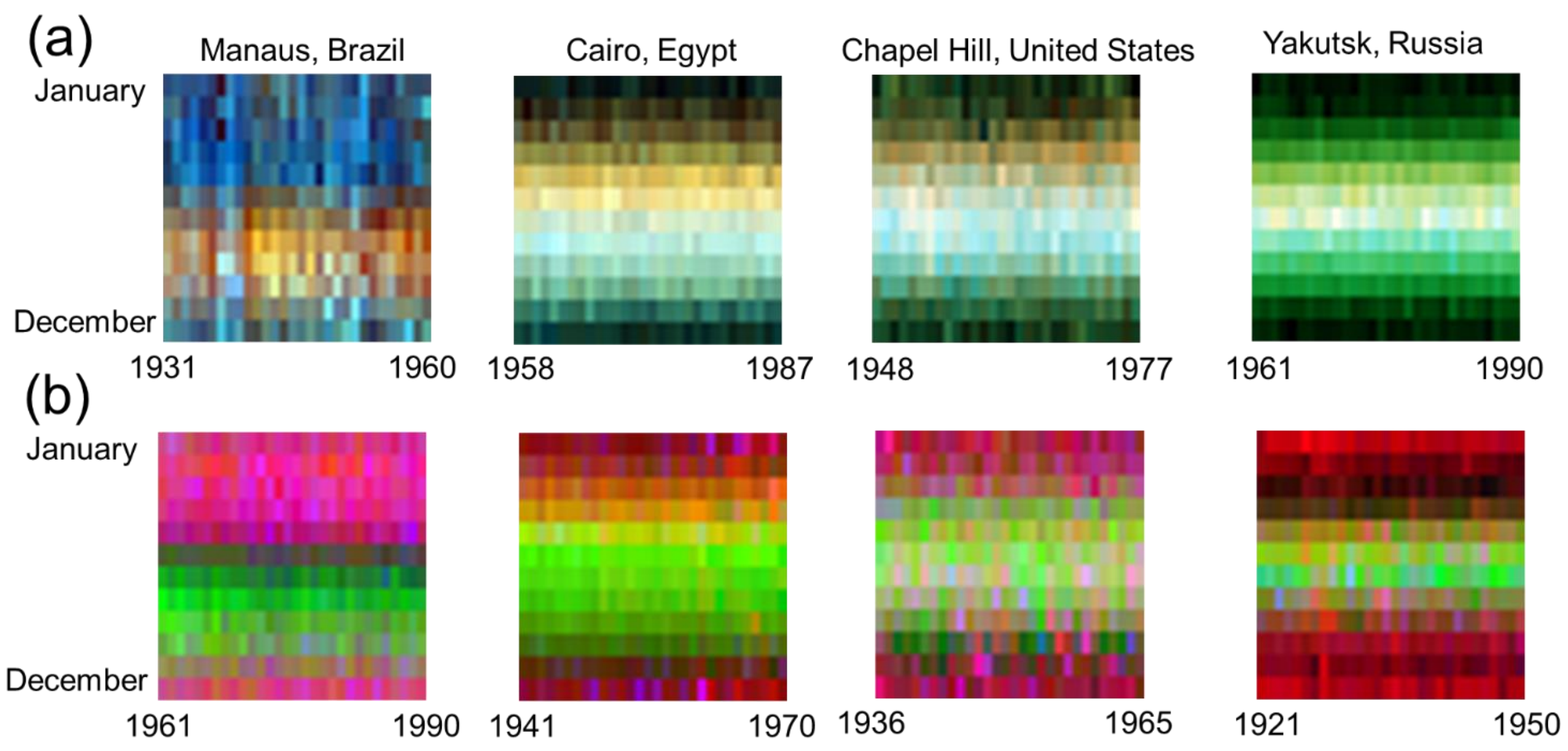

**Fig. 1.**

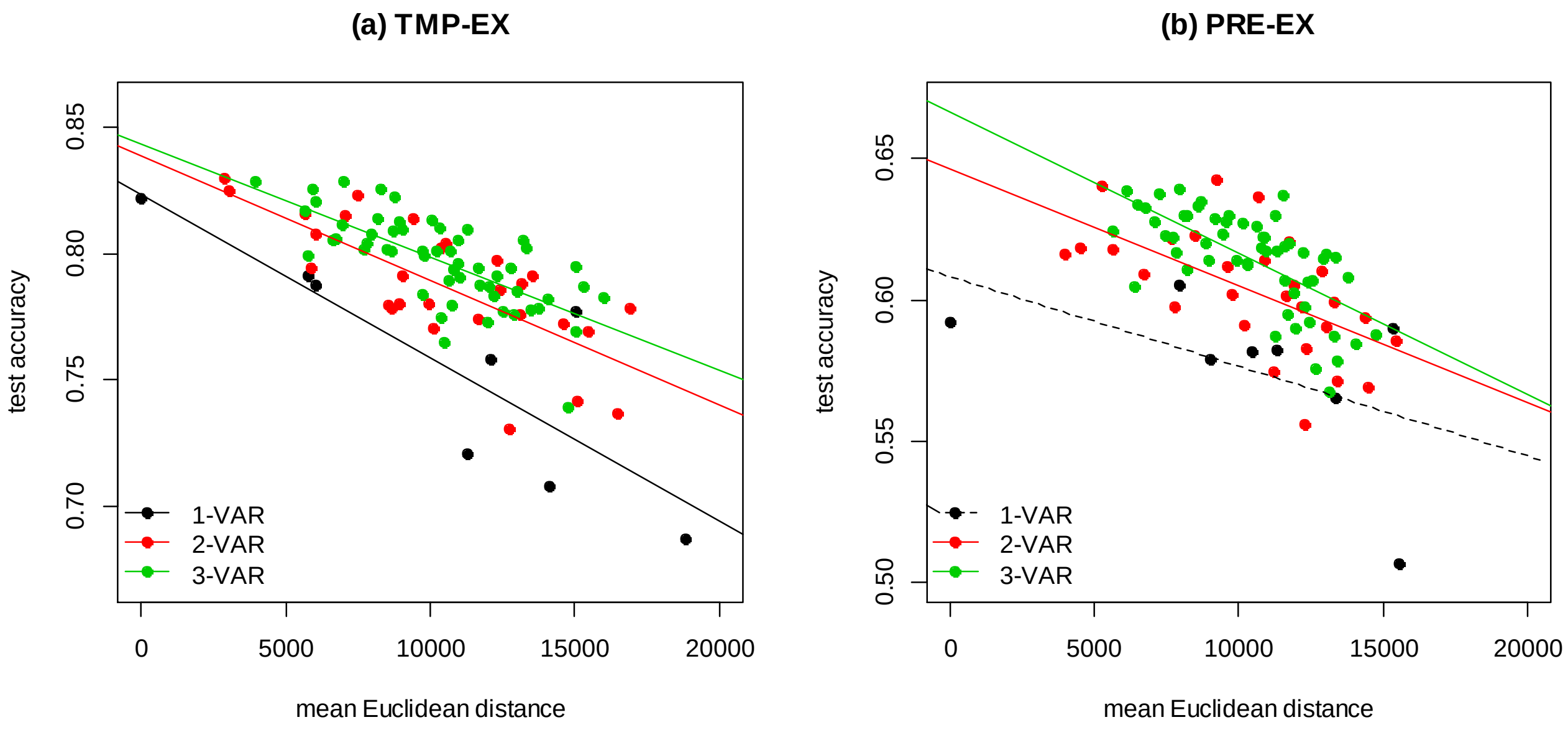

**Fig. 2.**

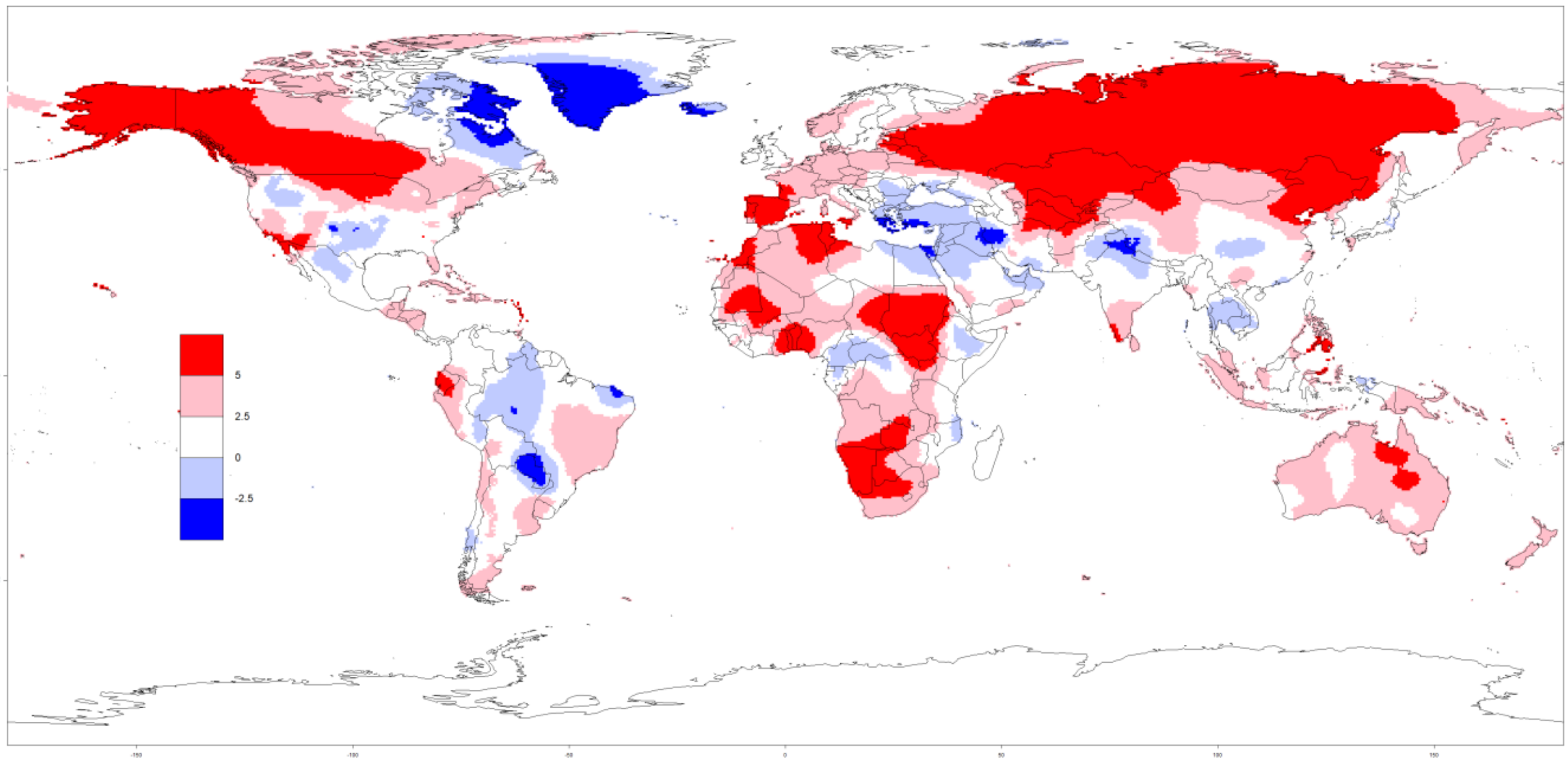

**(a)**

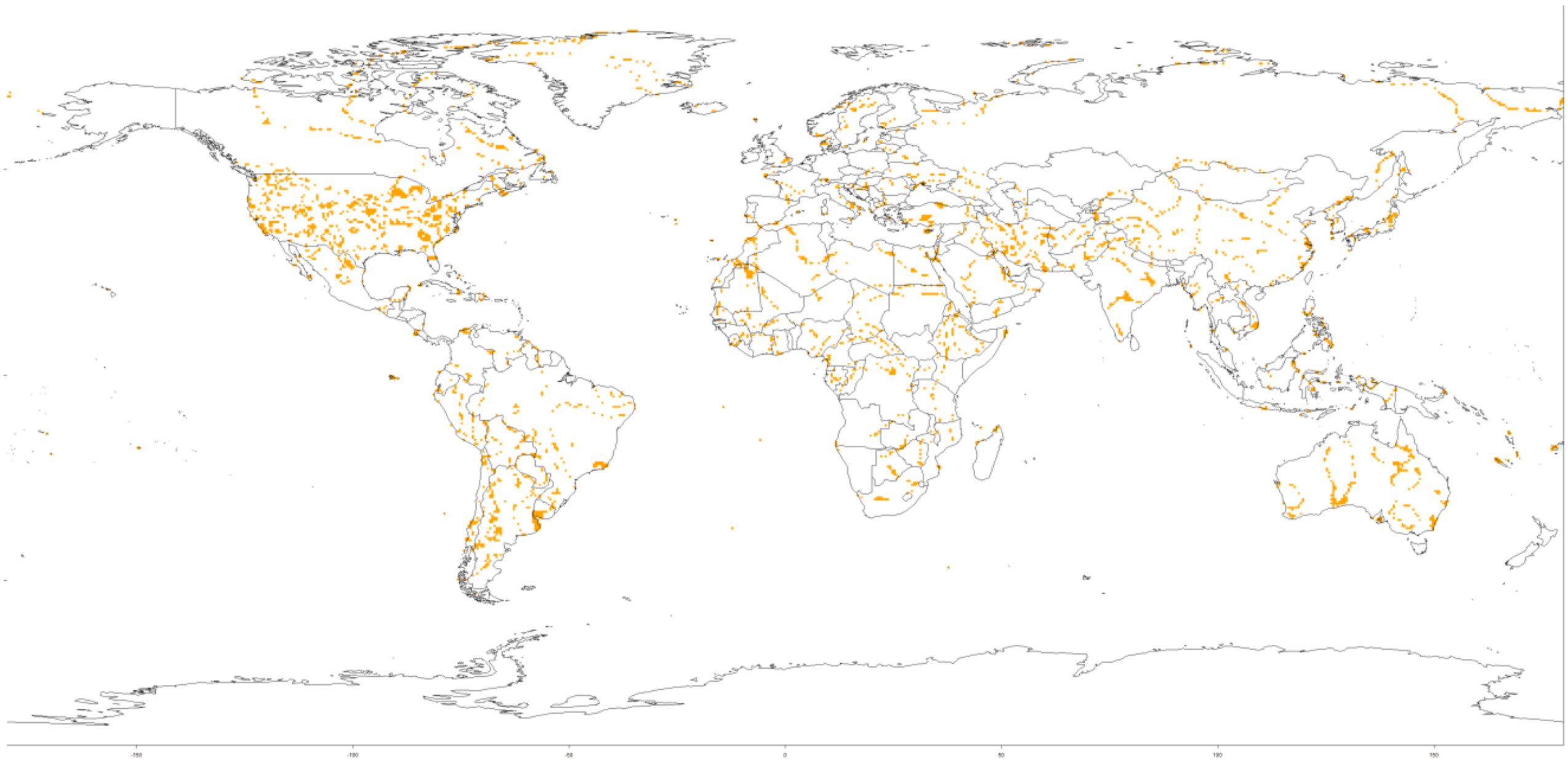

**(b)**

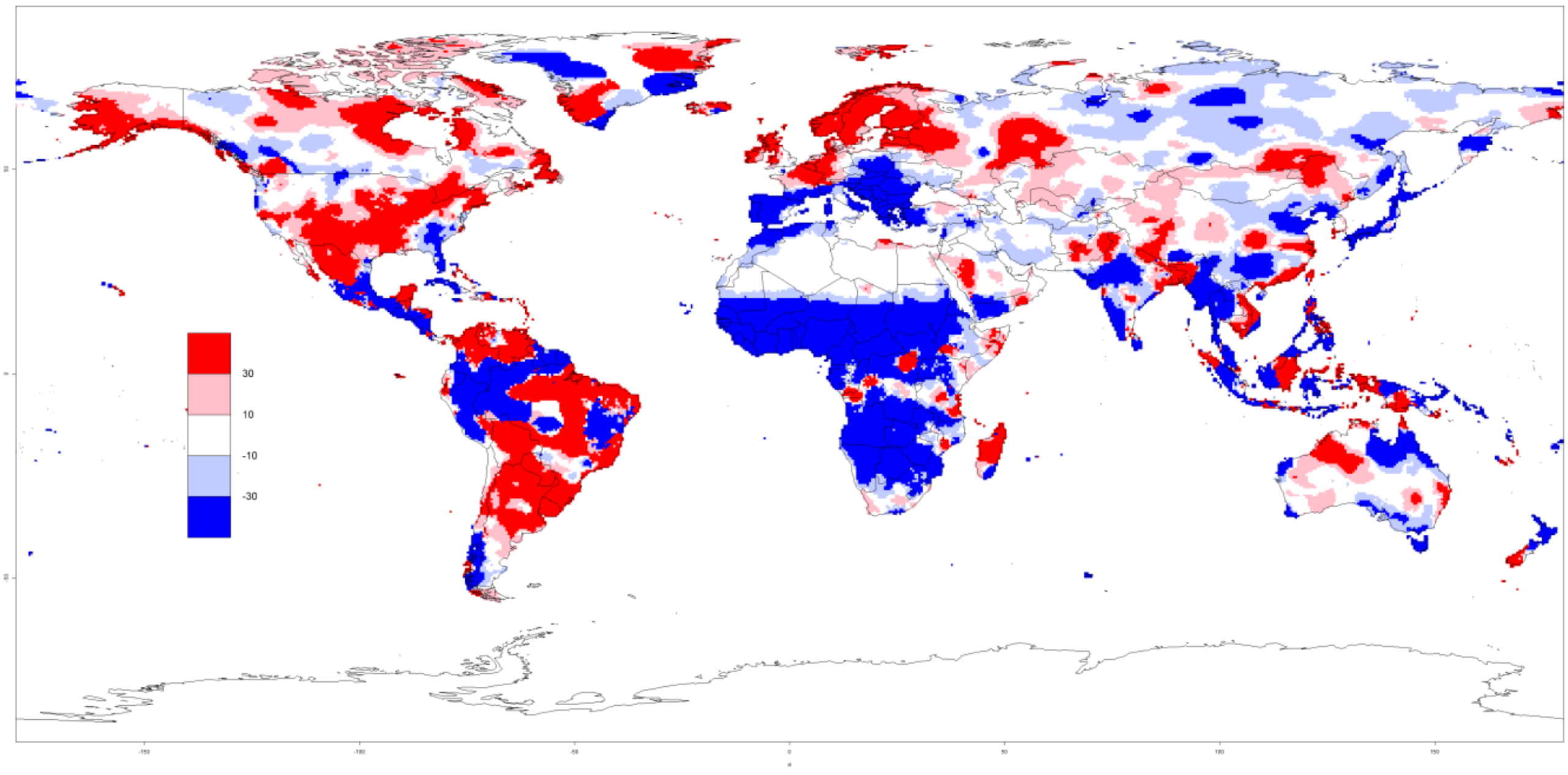

**(c)**

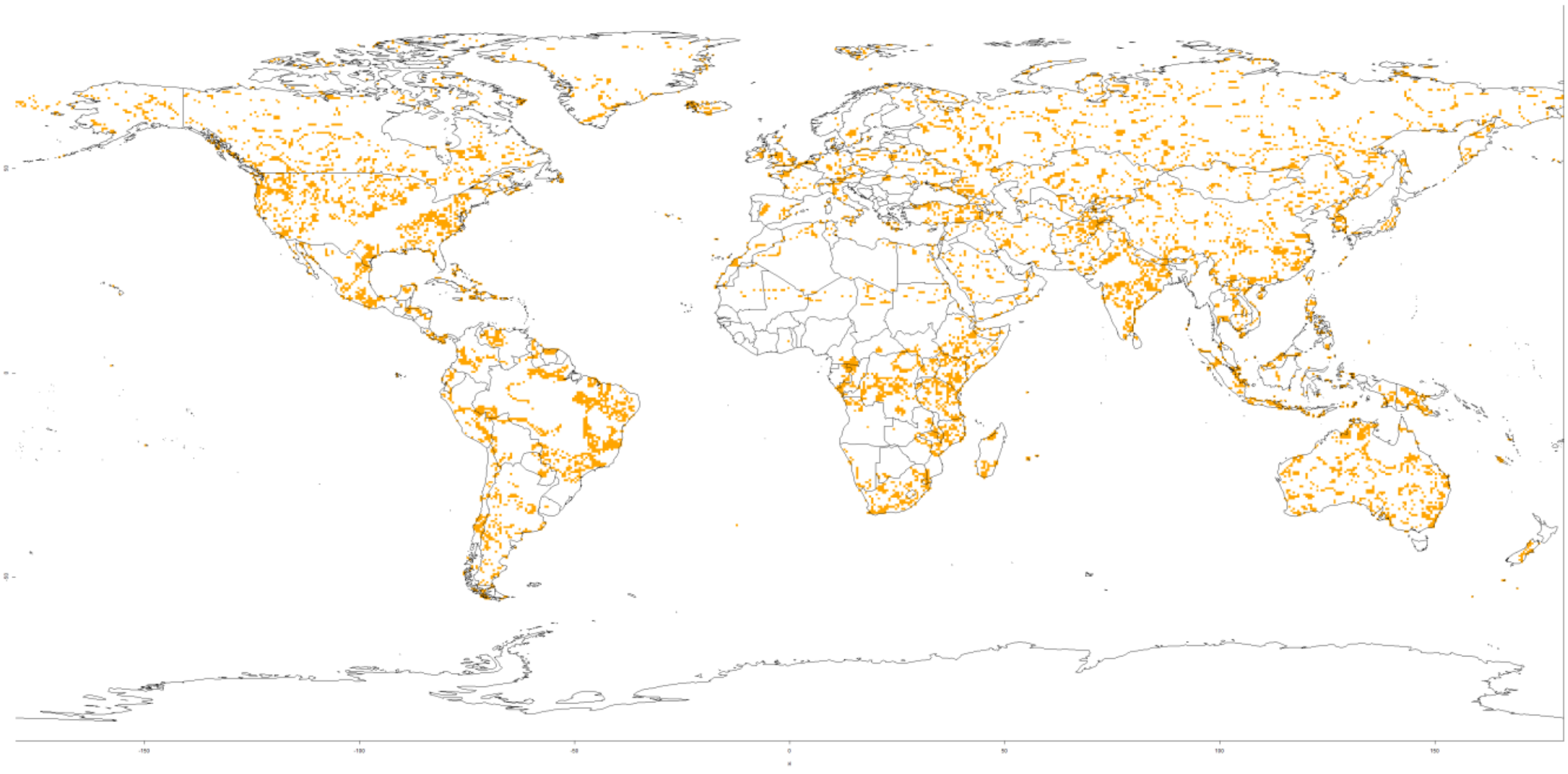

**(d)**

**Fig. 3.**

**Supplementary Table 1.** Summary of combinatorial experiments and resulting test accuracies.

| model No. | cld | dtr | frs | pet | pre | tmp | vap | wet | test accuracy (%) | |
|---|---|---|---|---|---|---|---|---|---|---|
| | | | | | | | | | TMP-EX | PRE-EX |
| #1 | cld | - | - | - | - | - | - | - | 77.72 | 58.15 |
| #2 | - | dtr | - | - | - | - | - | - | 75.83 | 59.01 |
| #3 | - | - | frs | - | - | - | - | - | 68.74 | 50.69 |
| #4 | - | - | - | pet | - | - | - | - | 79.1 | 56.51 |
| #5 | - | - | - | - | pre | - | - | - | 72.08 | 59.23 |
| #6 | - | - | - | - | - | tmp | - | - | 82.17 | 58.2 |
| #7 | - | - | - | - | - | - | vap | - | 78.74 | 57.87 |
| #8 | - | - | - | - | - | - | - | wet | 70.81 | 60.53 |
| #9 | cld | dtr | - | - | - | - | - | - | 79.13 | 61.02 |
| #10 | cld | - | frs | - | - | - | - | - | 77.85 | 59.03 |
| #11 | cld | - | - | pet | - | - | - | - | 80.2 | 60.52 |
| #12 | cld | - | - | - | pre | - | - | - | 78.84 | 64.04 |
| #13 | cld | - | - | - | - | tmp | - | - | 82.3 | 61.39 |
| #14 | cld | - | - | - | - | - | vap | - | 80.39 | 60.17 |
| #15 | cld | - | - | - | - | - | - | wet | 77.2 | 64.25 |
| #16 | - | dtr | frs | - | - | - | - | - | 76.91 | 58.56 |
| #17 | - | dtr | - | pet | - | - | - | - | 78.02 | 59.35 |
| #18 | - | dtr | - | - | pre | - | - | - | 77.44 | 62.16 |
| #19 | - | dtr | - | - | - | tmp | - | - | 80.75 | 59.92 |
| #20 | - | dtr | - | - | - | - | vap | - | 79.15 | 59.74 |
| #21 | - | dtr | - | - | - | - | - | wet | 77.61 | 60.11 |
| #22 | - | - | frs | pet | - | - | - | - | 79.73 | 56.92 |
| #23 | - | - | frs | - | pre | - | - | - | 74.15 | 59.73 |
| #24 | - | - | frs | - | - | tmp | - | - | 81.39 | 57.12 |
| #25 | - | - | frs | - | - | - | vap | - | 78.59 | 55.58 |
| #26 | - | - | frs | - | - | - | - | wet | 73.69 | 62.05 |
| #27 | - | - | - | pet | pre | - | - | - | 77.97 | 60.92 |
| #28 | - | - | - | pet | - | tmp | - | - | 82.96 | 58.3 |
| #29 | - | - | - | pet | - | - | vap | - | 79.45 | 57.46 |
| #30 | - | - | - | pet | - | - | - | wet | 78.05 | 63.63 |
| #31 | - | - | - | - | pre | tmp | - | - | 81.6 | 61.79 |
| #32 | - | - | - | - | pre | - | vap | - | 77.84 | 61.84 |
| #33 | - | - | - | - | pre | - | - | wet | 73.08 | 61.64 |
| #34 | - | - | - | - | - | tmp | vap | - | 82.5 | 59.11 |
| #35 | - | - | - | - | - | tmp | - | wet | 81.51 | 61.16 |
| #36 | - | - | - | - | - | - | vap | wet | 77.06 | 62.28 |
| #37 | cld | dtr | frs | - | - | - | - | - | 78.68 | 60.79 |
| #38 | cld | dtr | - | pet | - | - | - | - | 79.62 | 61.6 |
| #39 | cld | dtr | - | - | pre | - | - | - | 79.42 | 63.3 |
| #40 | cld | dtr | - | - | - | tmp | - | - | 80.94 | 60.65 |
| #41 | cld | dtr | - | - | - | - | vap | - | 79.04 | 61.87 |
| #42 | cld | dtr | - | - | - | - | - | wet | 77.86 | 62.99 |
| #43 | cld | - | frs | pet | - | - | - | - | 80.53 | 56.76 |
| #44 | cld | - | frs | - | pre | - | - | - | 79.49 | 63.5 |
| #45 | cld | - | frs | - | - | tmp | - | - | 80.99 | 59.19 |
| #46 | cld | - | frs | - | - | - | vap | - | 80.21 | 59.5 |
| #47 | cld | - | frs | - | - | - | - | wet | 78.24 | 61.72 |
| #48 | cld | - | - | pet | pre | - | - | - | 80.12 | 63.89 |
| #49 | cld | - | - | pet | - | tmp | - | - | 81.13 | 62.01 |
| #50 | cld | - | - | pet | - | - | vap | - | 81.26 | 61.74 |
| #51 | cld | - | - | pet | - | - | - | wet | 79.45 | 62.61 |
| #52 | cld | - | - | - | pre | tmp | - | - | 82.27 | 63.75 |
| #53 | cld | - | - | - | pre | - | vap | - | 79.4 | 63.39 |
| #54 | cld | - | - | - | pre | - | - | wet | 77.8 | 63.86 |
| #55 | cld | - | - | - | - | tmp | vap | - | 82.84 | 61.24 |
| #56 | cld | - | - | - | - | tmp | - | wet | 80.09 | 61.39 |
| #57 | cld | - | - | - | - | - | vap | wet | 78.76 | 62.85 |
| #58 | - | dtr | frs | pet | - | - | - | - | 78.36 | 58.77 |
| #59 | - | dtr | frs | - | pre | - | - | - | 78.22 | 61.26 |
| #60 | - | dtr | frs | - | - | tmp | - | - | 81.01 | 58.46 |
| #61 | - | dtr | frs | - | - | - | vap | - | 79.12 | 58.72 |
| #62 | - | dtr | frs | - | - | - | - | wet | 76.9 | 61.46 |
| #63 | - | dtr | - | pet | pre | - | - | - | 78.37 | 62.74 |
| #64 | - | dtr | - | pet | - | tmp | - | - | 82.58 | 61.49 |
| #65 | - | dtr | - | pet | - | - | vap | - | 80.78 | 60.69 |
| #66 | - | dtr | - | pet | - | - | - | wet | 78.94 | 61.69 |
| #67 | - | dtr | - | - | pre | tmp | - | - | 80.43 | 62 |
| #68 | - | dtr | - | - | pre | - | vap | - | 79.93 | 62.98 |
| #69 | - | dtr | - | - | pre | - | - | wet | 77.7 | 62.23 |
| #70 | - | dtr | - | - | - | tmp | vap | - | 82.07 | 60.26 |
| #71 | - | dtr | - | - | - | tmp | - | wet | 80.89 | 63.71 |
| #72 | - | dtr | - | - | - | - | vap | wet | 77.97 | 61.81 |
| #73 | - | - | frs | pet | pre | - | - | - | 77.29 | 63.01 |
| #74 | - | - | frs | pet | - | tmp | - | - | 81.41 | 57.86 |
| #75 | - | - | frs | pet | - | - | vap | - | 80.1 | 57.59 |

| | | | | | | | | | | | |
|---|---|---|---|---|---|---|---|---|---|---|---|
| #76 | - | - | frs | pet | - | - | - | wet | 77.57 | 59.77 |
| #77 | - | - | frs | - | pre | tmp | - | - | 81.33 | 61.38 |
| #78 | - | - | frs | - | pre | - | vap | - | 78.69 | 61.09 |
| #79 | - | - | frs | - | pre | - | - | wet | 73.9 | 61.68 |
| #80 | - | - | frs | - | - | tmp | vap | - | 82.56 | 58.99 |
| #81 | - | - | frs | - | - | tmp | - | wet | 80.54 | 60.69 |
| #82 | - | - | frs | - | - | - | vap | wet | 78.53 | 62.22 |
| #83 | - | - | - | pet | pre | tmp | - | - | 81.68 | 62.97 |
| #84 | - | - | - | pet | pre | - | vap | - | 80.16 | 62.29 |
| #85 | - | - | - | pet | pre | - | - | wet | 77.45 | 62.76 |
| #86 | - | - | - | pet | - | tmp | vap | - | 82.87 | 58.7 |
| #87 | - | - | - | pet | - | tmp | - | wet | 80.53 | 62.2 |
| #88 | - | - | - | pet | - | - | vap | wet | 80.09 | 62.7 |
| #89 | - | - | - | - | pre | tmp | vap | - | 79.9 | 63.28 |
| #90 | - | - | - | - | pre | tmp | - | wet | 80.15 | 60.49 |
| #91 | - | - | - | - | pre | - | vap | wet | 76.47 | 62.45 |
| #92 | - | - | - | - | - | tmp | vap | wet | 80.57 | 62.31 |

**Supplementary Fig. 1**. Euclidean distances among the eight climatic variables.

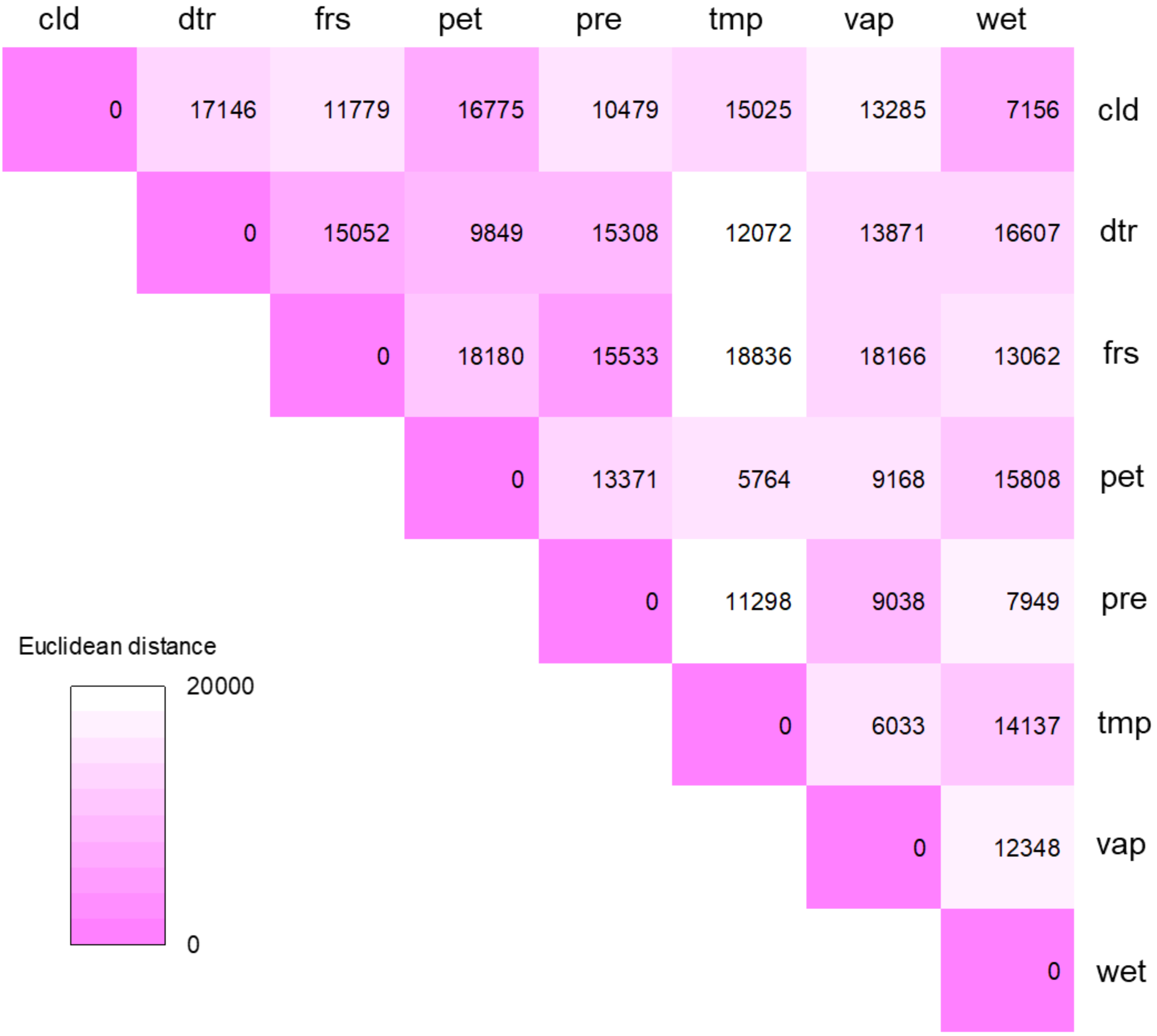

**Supplementary Fig. 2.** LeNet [18] employed in this study.

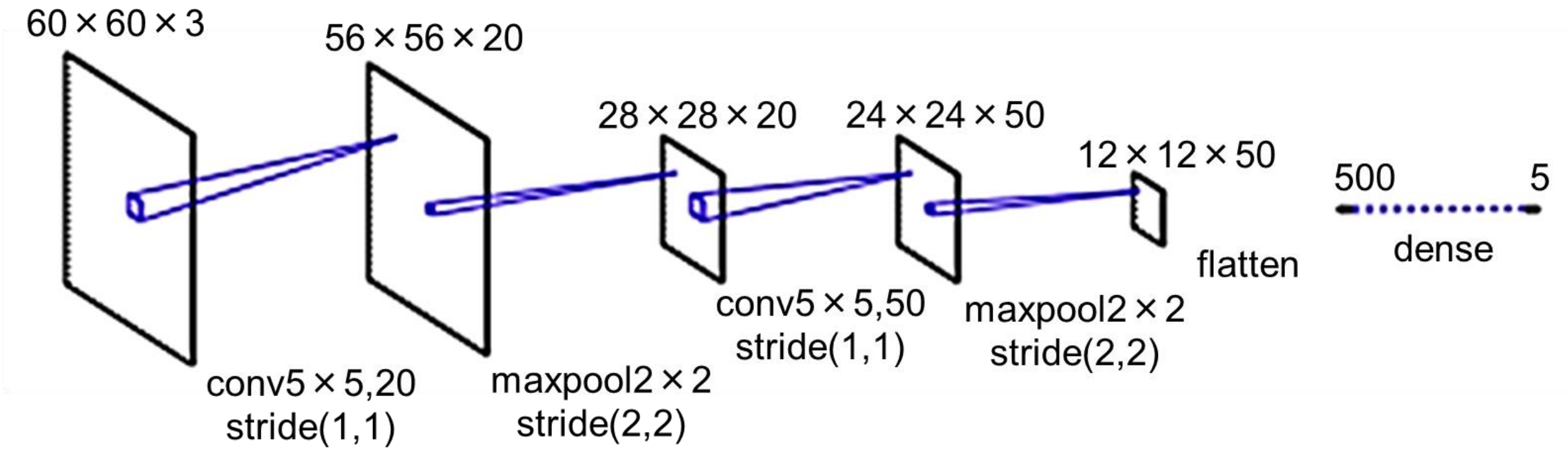

**Supplementary Fig. 3**. Samples of DIGITS 6.1 outputs for high-performance models with images ~1.5 million. Accuracy and loss for training and validation: (a) TMP-EX #86 and (b) PRE-EX #48. Number of images used for training by categories: (c) TMP-EX #86 and (d) PRE-EX #48.

(a)

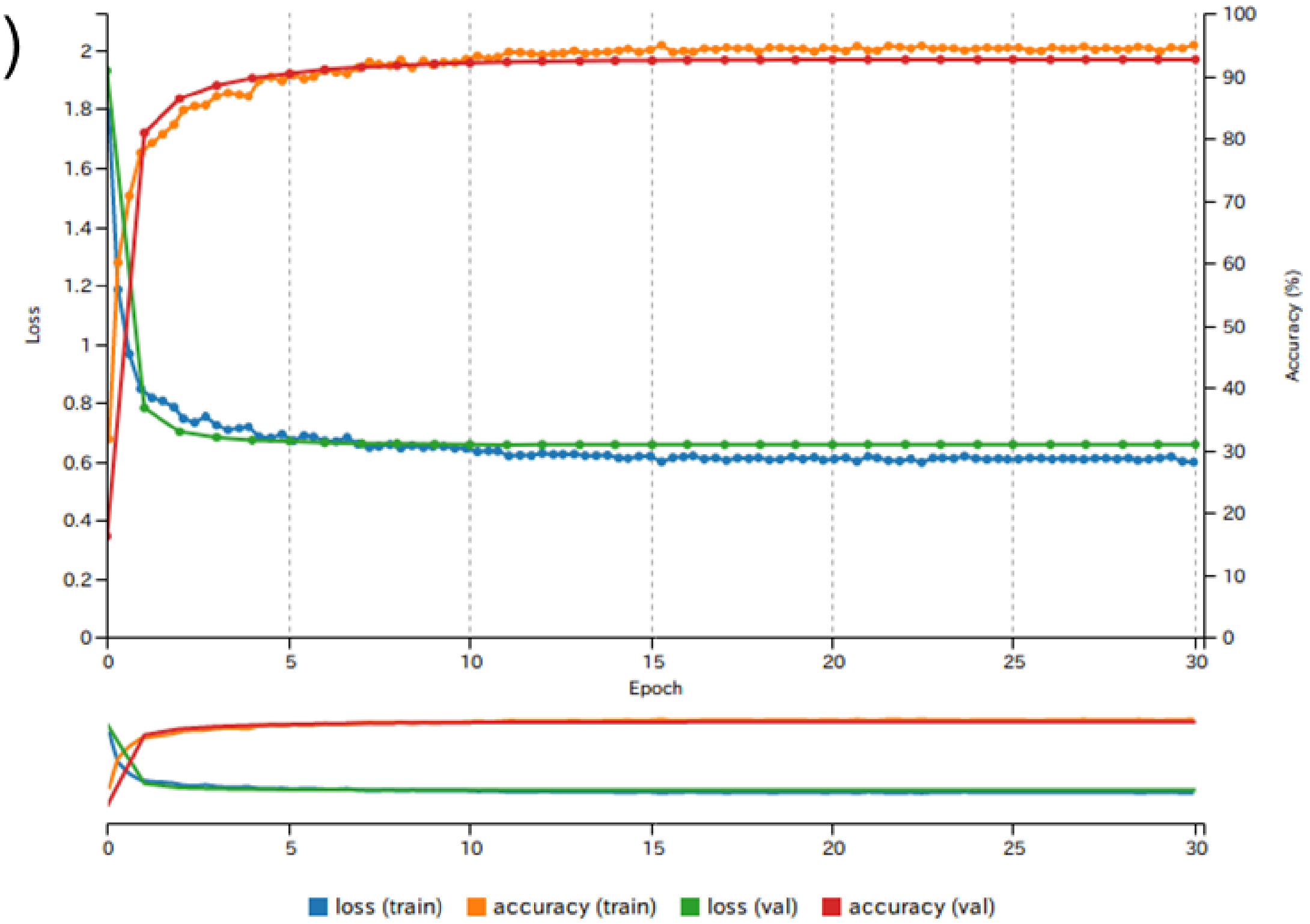

(b)

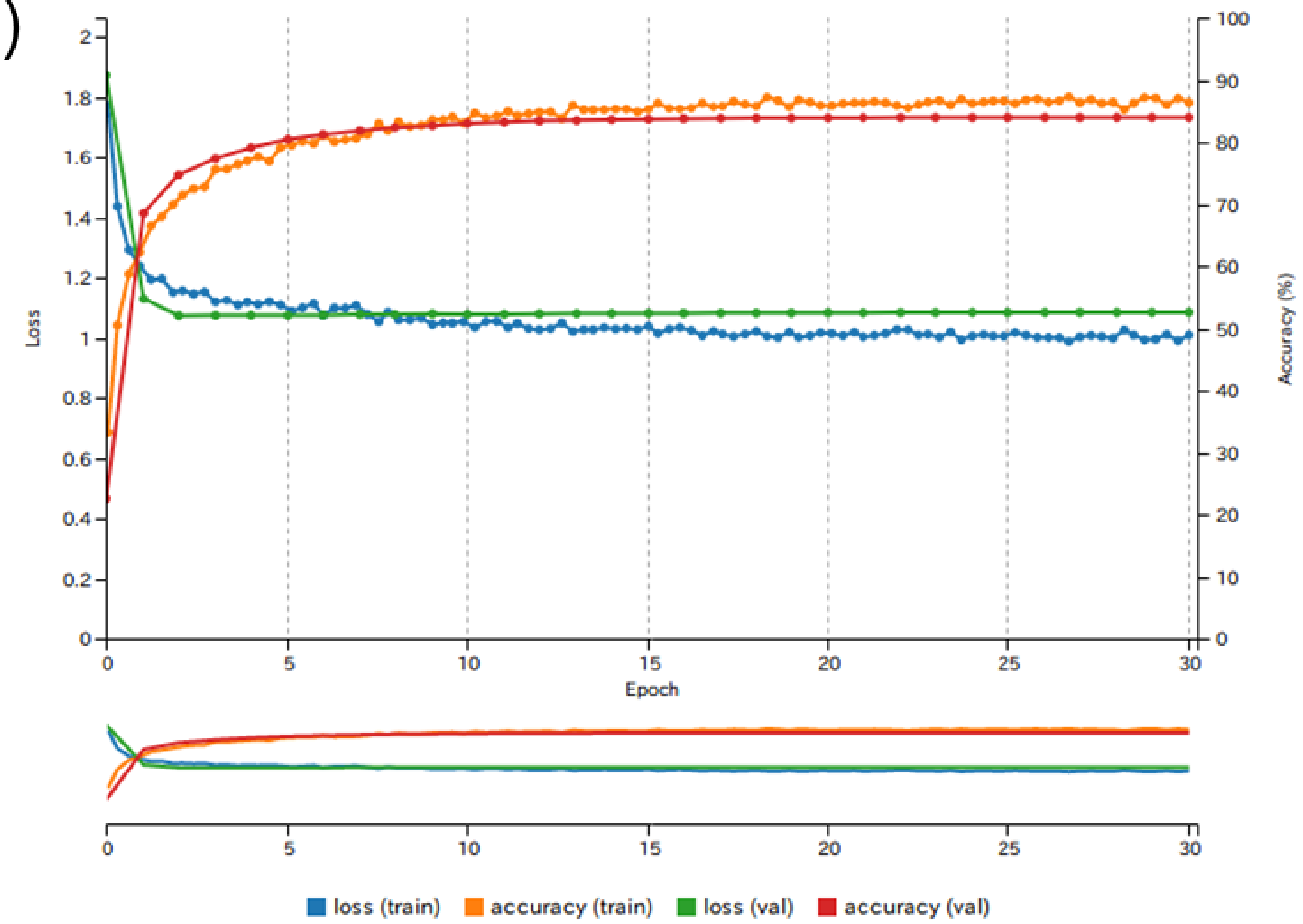

(c)

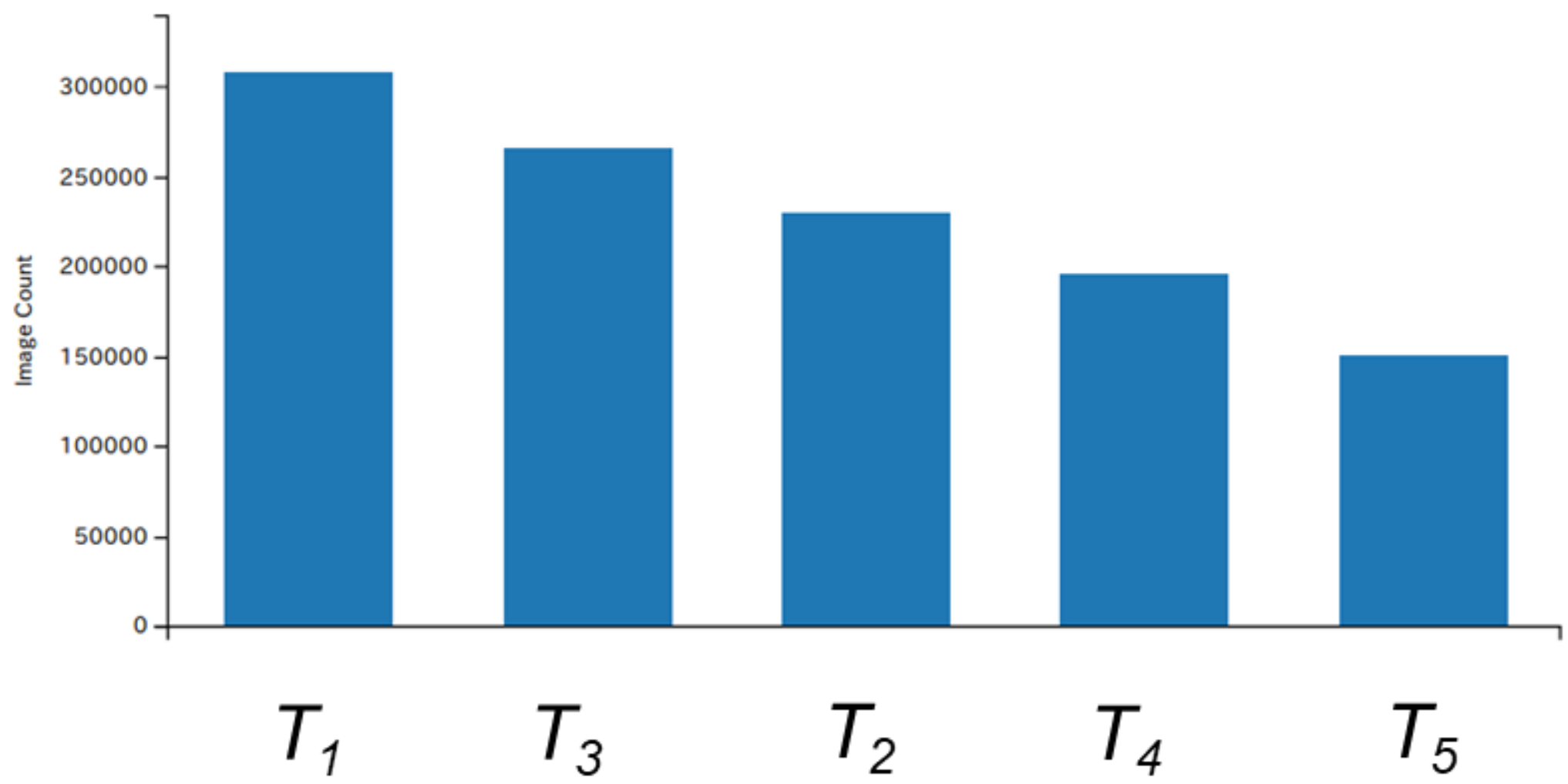

(d)

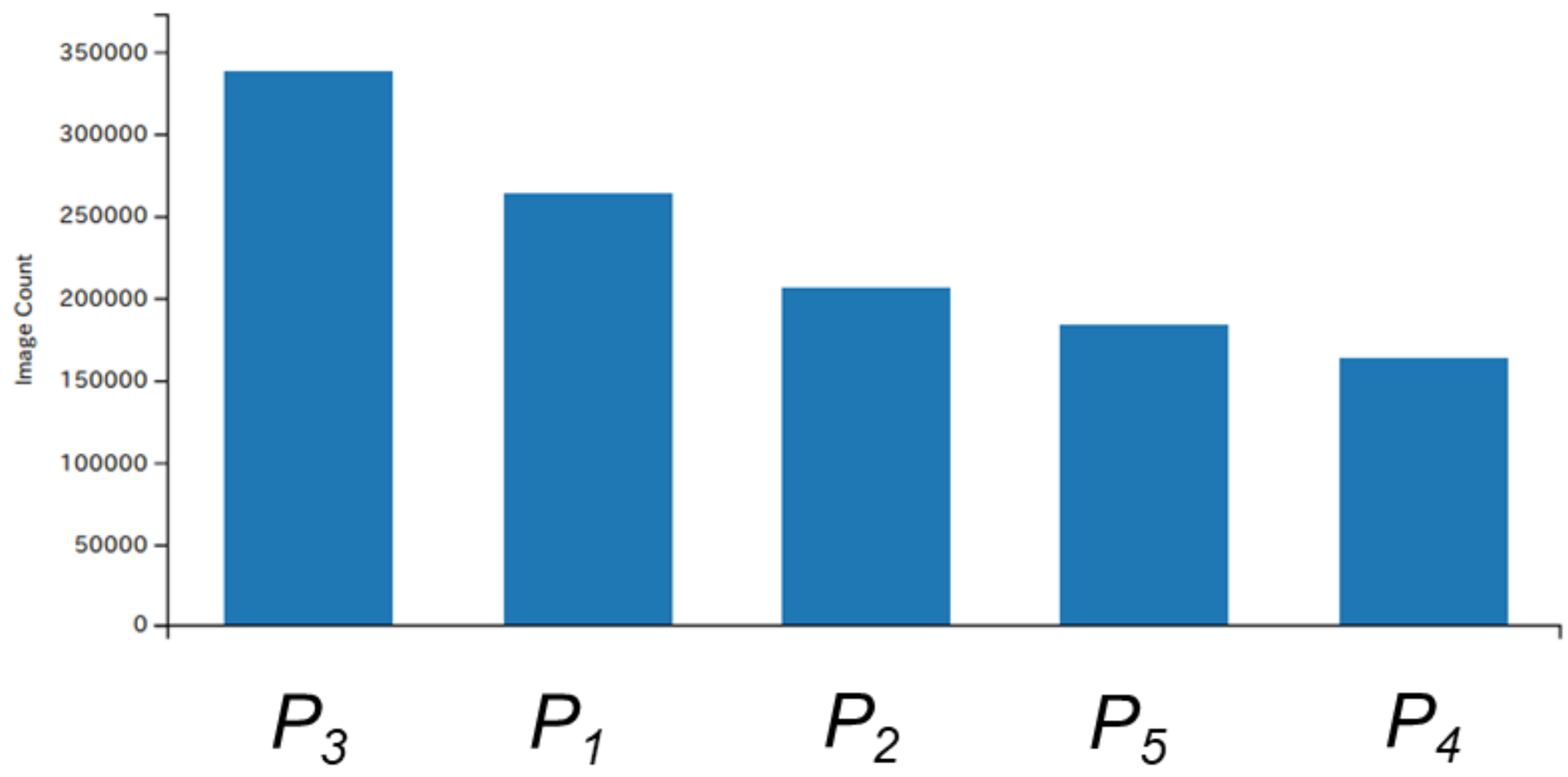